\documentclass{article}

    \usepackage[final]{neurips_2025}

\usepackage{atbegshi}

 \usepackage{amsmath}
\usepackage{makecell}
\usepackage{multirow}
\usepackage{graphicx} 
\usepackage{subcaption}
\usepackage{enumitem} 
\usepackage{booktabs} 
\usepackage{longtable} 
\usepackage[utf8]{inputenc} 
\usepackage[T1]{fontenc}    
\usepackage{hyperref}       
\usepackage{url}            
\usepackage{booktabs}       
\usepackage{amsfonts}       
\usepackage{nicefrac}       
\usepackage{microtype}      
\usepackage{xcolor}         
\usepackage{longtable}
\usepackage{array}
\usepackage{float}
\usepackage[table]{xcolor}
\usepackage{pgfplotstable}
\usepackage{booktabs}
\usepackage{colortbl}
\usepackage{caption}
\usepackage{geometry}
\usepackage{xcolor}
\usepackage{pgf}
\usepackage{makecell} 
\usepackage[table]{xcolor}
\usepackage{booktabs}
\usepackage{graphicx}
\usepackage{geometry}

\title{Out-of-Distribution Generalization in Climate-Aware Crop Yield Prediction with Earth Observation Data}

\author{%
  Aditya Chakravarty \\
  Independent Research \\
  San Francisco, CA \\
  \texttt{chakravarty.aditya28@gmail.com} \\
}

\AtBeginShipout{%
  \AtBeginShipoutFirst{%
    \vspace*{2mm}
    \par
    \centering
    {\normalsize Accepted in ICCV 2025 Workshop on Sustainability with Earth observation and AI}
    \vspace*{2mm}
  }%
}

\begin{document}

\maketitle

\section{Introduction and Related Work}

Climate change is increasingly destabilizing agricultural systems worldwide, with growing evidence of yield loss due to climate variability \citep{IPCC2021, Zhao2017, OrtizBobea2018}. Accurate crop yield forecasting has become critical for ensuring food security and sustainable planning under these non-stationary conditions \citep{Houghton1990, Leng2020}. Deep learning models have improved performance by capturing spatio-temporal dependencies in satellite and weather data \citep{khaki2019cnnrnn, Tseng2021}. However, most models remain untested under true out-of-distribution (OOD) conditions across geographies and time.Early neural network models for crop yield prediction demonstrated promising results, outperforming conventional regression techniques \citep{Drummond2003, Liu2001}. Among the meteorological and environmental data-based approaches, a key advancement was the CNN-RNN framework, which integrates multi-year meteorological and environmental data to improve yield forecasts \citep{khaki2019cnnrnn, khaki2021yieldnet}. This method established the importance of historical weather data, demonstrating that using multi-year sequences of climate variables significantly enhances prediction accuracy.

Building upon CNN-RNN architectures, newer methods incorporate graph neural networks (GNNs) to model geographical dependencies. The GNN-RNN model extends CNN-RNN by incorporating spatial relationships among counties, enabling the model to leverage information from neighboring regions to refine yield predictions \citep{fan2022gnnrnn} using long-term meteorological data. This method has shown improvements over CNN-RNN models in various evaluations, demonstrating the benefits of integrating spatial context into deep learning frameworks. Prior models fail to generalize across regions and years—an essential requirement for real-world deployment. In this work, we benchmark two state-of-the-art models—GNN-RNN \citep{fan2022gnnrnn} and MMST-ViT \citep{lin2023mmstvit}—under realistic spatio-temporal distribution shifts using the large-scale, publicly available CropNet dataset \citep{lin2024cropnet}. This work aims to identify geographic regions with stable transfer dynamics under climate variability and evaluate modeling approaches that best support robust cross-region generalization for climate-aware crop yield prediction.

\section{Dataset and Methods}
\begin{figure}[h]
    \centering
    \begin{minipage}[t]{0.48\linewidth}
        \centering
        \includegraphics[width=\linewidth]{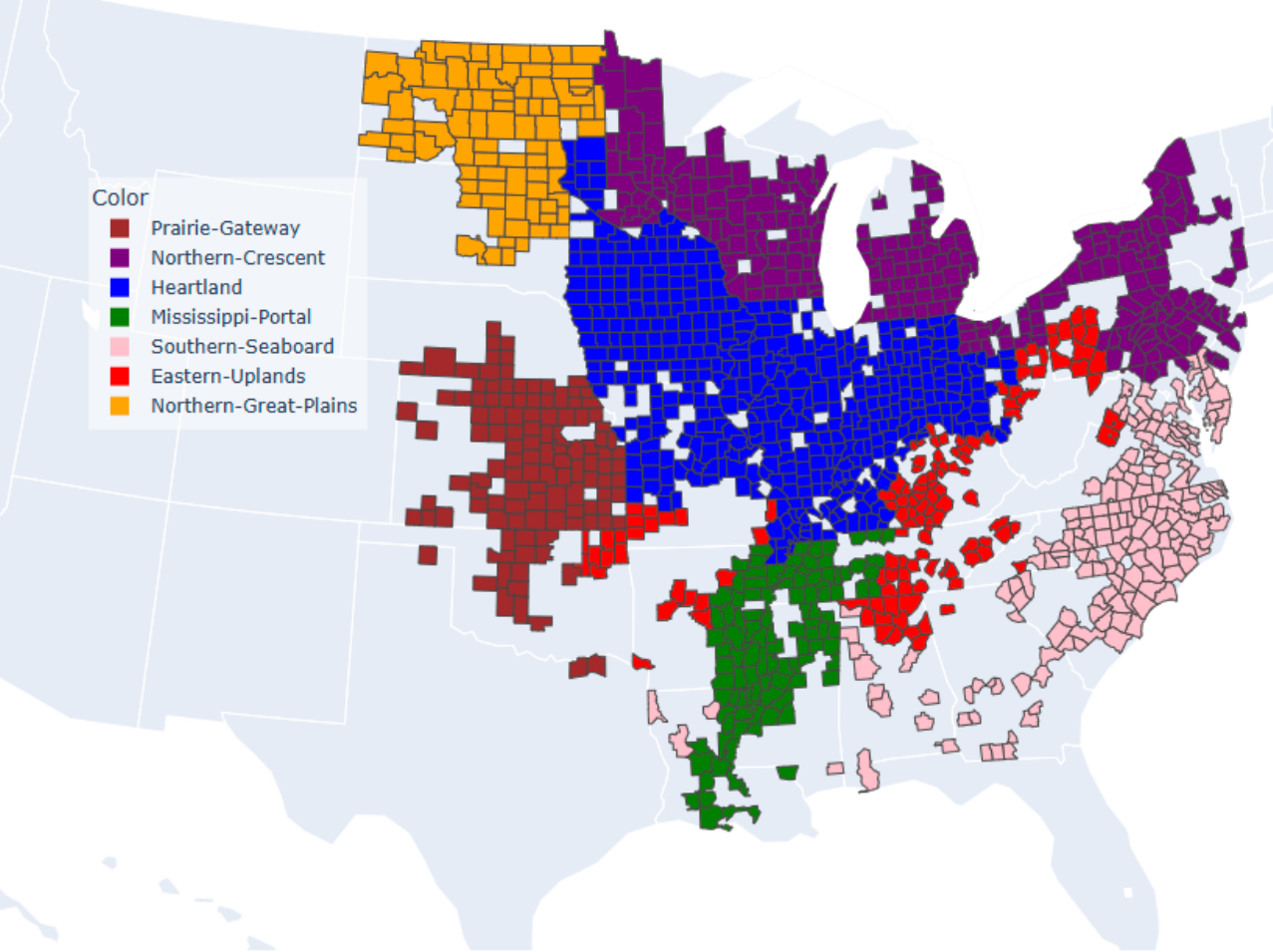}
        \caption{USDA Farm Resource Regions across 1,200 counties.}
        \label{fig:study_regions}
    \end{minipage}
    \hfill
    \begin{minipage}[t]{0.48\linewidth}
        \centering
        \includegraphics[width=\linewidth]{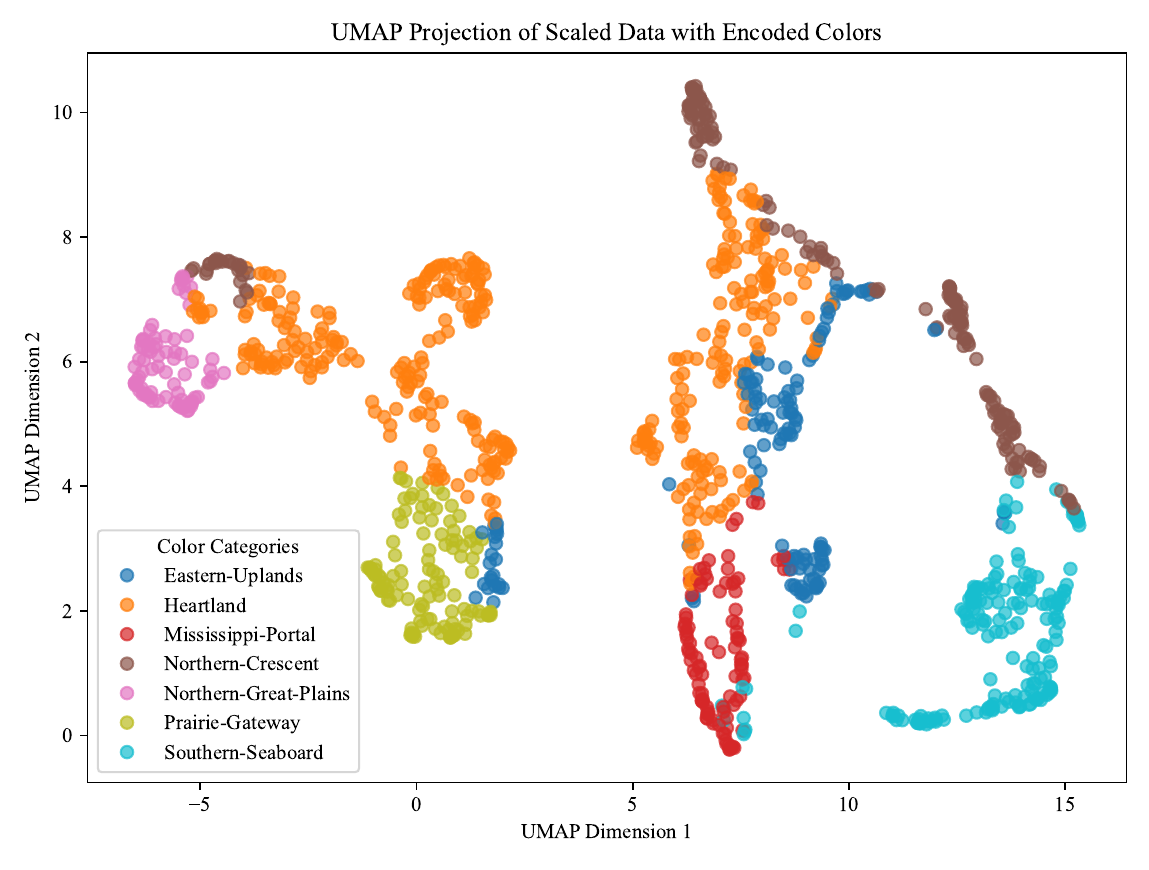}
        \caption{UMAP visualization of weekly weather embeddings (2017–2022), colored by USDA Farm Resource Regions. Clustering confirms that FRRs provide a meaningful partitioning of agricultural zones.}
        \label{fig:umap_embeds}
    \end{minipage}
\end{figure}

We used the CropNet dataset \citep[see][\footnotemark]{lin2024cropnet}, which is a large-scale, publicly available, multi-modal dataset specifically designed for climate change-aware crop yield predictions across the contiguous United States from 2017 to 2022. The CropNet dataset provides preprocessed Sentinel-2 imagery at 40m spatial resolution with a 14 day revisit cycle, optimized for agricultural monitoring across 2291 U.S. counties. Cloud coverage is limited to $\leq 20$\% using the Sentinel Hub API, and only select spectral bands (AG and NDVI) are retained (Figure~\ref{fig:study_regions}). This structured image processing pipeline supports robust tracking of seasonal crop dynamics critical for sustainable yield modeling.
\footnotetext{\url{https://huggingface.co/datasets/CropNet/CropNet}}
We define seven USDA Farm Resource Regions \citep{Heimlich2000farm, spangler2020past} as scientifically valid clusters for evaluating generalization (Figure~\ref{fig:umap_embeds}). We perform: (i) leave-one-cluster-out (LCO) CV and (ii) realistic year-ahead transfer with 3-to-1 train-test splits. Figure~\ref{fig:study_regions} shows the region map. GNN-RNN integrates LSTM over multi-year weather data with spatial message passing. MMST-ViT uses attention over fused weather and satellite inputs. Both are tuned via LCO and tested on 2022 data. The GNN-RNN model processes multi-year county-level weather data using CNNs and GNNs to capture temporal and spatial dependencies, which are then fed into an RNN to predict annual crop yields.

\begin{figure}[h]
    \centering
    \includegraphics[width=0.999\linewidth]{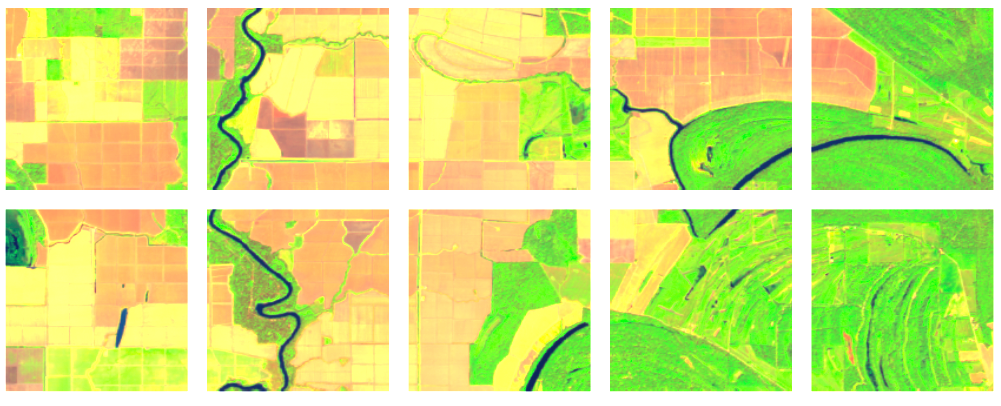}
    \caption{Sample Sentinel-2 image patches from the CropNet dataset ($\leq$20\% cloud cover), highlighting diverse crop patterns and landscapes across U.S. counties.}
    \label{fig:study_regions}
\end{figure}

\subsection{Cross-Validation, Ablation, and Real-World Scenarios}

Spatio-temporal generalization remained challenging: while year-ahead predictions showed moderate degradation, leave-one-region-out (LORO) settings led to substantial performance drops, often with negative $R^2$ and correlation values. Regions like Eastern Uplands (EU), Heartland (HL), and Northern Great Plains (NGP) showed relatively stable performance across models. The parameters tuned are embedding dimension (e), dropout (drop) and depth and aggregation type (n\_layers, agg, only in GNN). For soybean, HL and NGP achieved positive $R^2$ under GNN-RNN (\texttt{n\_layers=4}) and MMST-ViT (\texttt{e=512, drop=0}). For corn, NGP reached $R^2 \approx 0.45$ with MMST-ViT (\texttt{e=128, drop=0.5}).  GNN-RNN degraded with higher dropout, while deeper architectures helped in HL and NGP. MMST-ViT performed best with smaller embeddings and minimal regularization; larger sizes or stronger dropout led to severe overfitting in difficult regions like Prairie Gateway and Southern Seaboard. Based on these insights, we used: GNN-RNN with \texttt{n\_layers=4, dropout=0, agg=mean/pool}, and MMST-ViT with \texttt{e=128, drop=0}. Table~\ref{tab:ood_scenarios} defines OOD difficulty levels based on LCO results and cluster similarity.

\begin{table}[htbp]
    \caption{Real-world scenarios and corresponding USDA Farm Resource Region splits.}
    \label{tab:ood_scenarios}
    \centering
    \small
    \begin{tabular}{p{3.8cm} p{5cm} p{2.5cm}}
        \toprule
        \textbf{Scenario} & \textbf{Train Region} & \textbf{Test Region} \\
        \midrule
        Case 1 (Easy) & Prairie-Gateway + Heartland + Mississippi-Portal & Eastern-Uplands \\
        Case 2 (Medium) & Northern-Crescent + Prairie-Gateway + Northern-Great-Plains & Heartland \\
        Case 3 (Hard) & Prairie-Gateway + Southern-Seaboard + Mississippi-Portal & Northern-Great-Plains \\
        \bottomrule
    \end{tabular}
\end{table}

\section{Cross-region transferability and pairwise RMSE patterns}

\newcommand{\soycell}[2]{%
  \ifdim #1 pt < 8 pt \cellcolor{green!30}\textbf{#2}%
  \else\ifdim #1 pt < 11 pt \cellcolor{yellow!30}#2%
  \else\ifdim #1 pt < 14 pt \cellcolor{orange!30}#2%
  \else \cellcolor{red!30}#2%
  \fi\fi\fi
}

\newcommand{\corncell}[2]{%
  \ifdim #1 pt < 28 pt \cellcolor{green!30}\textbf{#2}%
  \else\ifdim #1 pt < 40 pt \cellcolor{yellow!30}#2%
  \else\ifdim #1 pt < 55 pt \cellcolor{orange!30}#2%
  \else \cellcolor{red!30}#2%
  \fi\fi\fi
}

\begin{table}[htbp]
\centering
\caption{RMSE for Soybean (left) and Corn (right) using GNN-RNN model; diagonal entries are bold represent year-ahead predictions for 2022 and colors indicate performance}
\resizebox{\textwidth}{!}{%
\begin{tabular}{>{\centering\arraybackslash}p{1cm} ccccccc @{\hskip 15pt} >{\centering\arraybackslash}p{1cm} ccccccc}

& EU & HL & MSP & NGP & NC & PG & SS & & EU & HL & MSP & NGP & NC & PG & SS \\
\textbf{Train$\backslash$Test} & \multicolumn{7}{c}{\textbf{Soybean}} & \textbf{Train$\backslash$Test} & \multicolumn{7}{c}{\textbf{Corn}} \\
\midrule
EU & \soycell{5.46}{5.46} & \soycell{7.24}{7.24} & \soycell{11.55}{11.55} & \soycell{9.01}{9.01} & \soycell{9.63}{9.63} & \soycell{21.42}{21.42} & \soycell{9.70}{9.70}
   & EU & \corncell{26.69}{26.69} & \corncell{38.92}{38.92} & \corncell{43.55}{43.55} & \corncell{32.86}{32.86} & \corncell{22.29}{22.29} & \corncell{97.42}{97.42} & \corncell{41.77}{41.77} \\
HL & \soycell{8.39}{8.39} & \soycell{6.15}{\textbf{6.15}} & \soycell{8.88}{8.88} & \soycell{7.73}{7.73} & \soycell{8.57}{8.57} & \soycell{19.30}{19.30} & \soycell{11.50}{11.50}
   & HL & \corncell{33.05}{33.05} & \corncell{22.67}{\textbf{22.67}} & \corncell{32.44}{32.44} & \corncell{32.20}{32.20} & \corncell{33.66}{33.66} & \corncell{56.21}{56.21} & \corncell{49.68}{49.68} \\
MSP & \soycell{17.22}{17.22} & \soycell{11.17}{11.17} & \soycell{6.09}{\textbf{6.09}} & \soycell{9.79}{9.79} & \soycell{9.72}{9.72} & \soycell{26.22}{26.22} & \soycell{10.59}{10.59}
    & MSP & \corncell{40.98}{40.98} & \corncell{23.48}{23.48} & \corncell{31.03}{\textbf{31.03}} & \corncell{30.87}{30.87} & \corncell{37.03}{37.03} & \corncell{53.04}{53.04} & \corncell{39.05}{39.05} \\
NGP & \soycell{8.94}{8.94} & \soycell{8.04}{8.04} & \soycell{9.70}{9.70} & \soycell{7.11}{\textbf{7.11}} & \soycell{11.88}{11.88} & \soycell{23.37}{23.37} & \soycell{12.48}{12.48}
    & NGP & \corncell{33.81}{33.81} & \corncell{25.83}{25.83} & \corncell{34.33}{34.33} & \corncell{21.65}{\textbf{21.65}} & \corncell{57.52}{57.52} & \corncell{51.49}{51.49} & \corncell{44.54}{44.54} \\
NC & \soycell{12.25}{12.25} & \soycell{12.39}{12.39} & \soycell{12.67}{12.67} & \soycell{10.29}{10.29} & \soycell{7.25}{\textbf{7.25}} & \soycell{28.23}{28.23} & \soycell{12.51}{12.51}
   & NC & \corncell{40.13}{40.13} & \corncell{41.67}{41.67} & \corncell{30.56}{30.56} & \corncell{34.64}{34.64} & \corncell{24.42}{\textbf{24.42}} & \corncell{92.17}{92.17} & \corncell{51.94}{51.94} \\
PG & \soycell{14.71}{14.71} & \soycell{11.43}{11.43} & \soycell{15.54}{15.54} & \soycell{14.62}{14.62} & \soycell{13.59}{13.59} & \soycell{11.71}{\textbf{11.71}} & \soycell{13.03}{13.03}
   & PG & \corncell{48.36}{48.36} & \corncell{33.31}{33.31} & \corncell{42.21}{42.21} & \corncell{41.59}{41.59} & \corncell{52.47}{52.47} & \corncell{42.94}{\textbf{42.94}} & \corncell{45.18}{45.18} \\
SS & \soycell{13.55}{13.55} & \soycell{10.69}{10.69} & \soycell{12.17}{12.17} & \soycell{12.86}{12.86} & \soycell{9.77}{9.77} & \soycell{11.20}{11.20} & \soycell{7.96}{\textbf{7.96}}
   & SS & \corncell{42.13}{42.13} & \corncell{29.91}{29.91} & \corncell{39.46}{39.46} & \corncell{41.47}{41.47} & \corncell{32.35}{32.35} & \corncell{53.41}{53.41} & \corncell{25.09}{\textbf{25.09}} \\
\end{tabular}%
}
\end{table}

\begin{table}[htbp]
\centering
\caption{RMSE for Soybean (left) and Corn (right) using MMSt-ViT model; diagonal entries are bold represent year-ahead predictions for 2022 and colors indicate performance}
\resizebox{\textwidth}{!}{%
\begin{tabular}{>{\centering\arraybackslash}p{1cm} ccccccc @{\hskip 15pt} >{\centering\arraybackslash}p{1cm} ccccccc}

& EU & HL & MSP & NC & NGP & PG & SS & & EU & HL & MSP & NC & NGP & PG & SS \\
\textbf{Train$\backslash$Test} & \multicolumn{7}{c}{\textbf{Soybean}} & \textbf{Train$\backslash$Test} & \multicolumn{7}{c}{\textbf{Corn}} \\
\midrule
EU & \soycell{8.93}{8.93} & \soycell{11.16}{11.16} & \soycell{9.69}{9.69} & \soycell{9.38}{9.38} & \soycell{14.51}{14.51} & \soycell{22.15}{22.15} & \soycell{15.25}{15.25}
   & EU & \corncell{26.06}{26.06} & \corncell{38.83}{38.83} & \corncell{35.82}{35.82} & \corncell{29.14}{29.14} & \corncell{34.29}{34.29} & \corncell{59.27}{59.27} & \corncell{45.90}{45.90} \\
HL & \soycell{11.93}{11.93} & \soycell{10.18}{\textbf{10.18}} & \soycell{11.89}{11.89} & \soycell{14.34}{14.34} & \soycell{25.01}{25.01} & \soycell{28.22}{28.22} & \soycell{20.58}{20.58}
   & HL & \corncell{43.70}{43.70} & \corncell{33.37}{\textbf{33.37}} & \corncell{52.20}{52.20} & \corncell{52.20}{52.20} & \corncell{52.38}{52.38} & \corncell{89.87}{89.87} & \corncell{69.84}{69.84} \\
MSP & \soycell{9.62}{9.62} & \soycell{11.91}{11.91} & \soycell{8.71}{\textbf{8.71}} & \soycell{9.72}{9.72} & \soycell{16.87}{16.87} & \soycell{23.15}{23.15} & \soycell{10.78}{10.78}
    & MSP & \corncell{35.21}{35.21} & \corncell{39.39}{39.39} & \corncell{41.32}{\textbf{41.32}} & \corncell{35.96}{35.96} & \corncell{49.69}{49.69} & \corncell{69.85}{69.85} & \corncell{57.16}{57.16} \\
NC & \soycell{11.34}{11.34} & \soycell{10.45}{10.45} & \soycell{10.13}{10.13} & \soycell{14.49}{\textbf{14.49}} & \soycell{19.58}{19.58} & \soycell{24.76}{24.76} & \soycell{14.26}{14.26}
    & NC & \corncell{36.32}{36.32} & \corncell{37.78}{37.78} & \corncell{45.47}{45.47} & \corncell{37.93}{\textbf{37.93}} & \corncell{40.94}{40.94} & \corncell{66.12}{66.12} & \corncell{45.27}{45.27} \\
NGP & \soycell{24.12}{24.12} & \soycell{25.63}{25.63} & \soycell{25.67}{25.67} & \soycell{16.59}{16.59} & \soycell{11.33}{\textbf{11.33}} & \soycell{17.76}{17.76} & \soycell{12.17}{12.17}
    & NGP & \corncell{44.55}{44.55} & \corncell{73.04}{73.04} & \corncell{60.21}{60.21} & \corncell{51.72}{51.72} & \corncell{43.61}{\textbf{43.61}} & \corncell{59.47}{59.47} & \corncell{36.88}{36.88} \\
PG & \soycell{12.23}{12.23} & \soycell{16.18}{16.18} & \soycell{15.87}{15.87} & \soycell{12.03}{12.03} & \soycell{15.37}{15.37} & \soycell{24.48}{\textbf{24.48}} & \soycell{14.48}{14.48}
   & PG & \corncell{38.69}{38.69} & \corncell{64.08}{64.08} & \corncell{60.93}{60.93} & \corncell{44.85}{44.85} & \corncell{41.15}{41.15} & \corncell{42.78}{\textbf{42.78}} & \corncell{46.92}{46.92} \\
SS & \soycell{13.55}{13.55} & \soycell{17.65}{17.65} & \soycell{10.55}{10.55} & \soycell{9.34}{9.34} & \soycell{8.49}{8.49} & \soycell{18.75}{18.75} & \soycell{7.15}{\textbf{7.15}}
   & SS & \corncell{30.22}{30.22} & \corncell{51.23}{51.23} & \corncell{33.48}{33.48} & \corncell{32.41}{32.41} & \corncell{30.57}{30.57} & \corncell{51.83}{51.83} & \corncell{35.48}{\textbf{35.48}} \\
\end{tabular}%
}
\end{table}

\begin{table}[htbp]
  \centering
    \small
  \caption{Training time comparison of MMST-ViT and GNN-RNN on a single RTX 4090 GPU. GNN-RNN achieves a $\sim$135$\times$ speedup over MMST-ViT.}
  \label{tab:training_time_comparison}
  \begin{tabular}{lccc}
    \toprule
    \textbf{Model} & \textbf{Pretraining Time} & \textbf{Fine-tuning Time} & \textbf{Total Training Time} \\
    \midrule
    MMST-ViT & 23 hours & 8.5 hours & 31.5 hours \\
    GNN-RNN & \multicolumn{2}{c}{--} & 14 minutes \\
    \bottomrule
  \end{tabular}
\end{table}

GNN-RNN consistently outperforms MMST-ViT across both crops in cross-region prediction. For soybean, HL, MSP, and NGP yield the lowest RMSEs, with HL$\rightarrow$MSP (8.88) and NGP$\rightarrow$HL (8.04) showing strong generalization. PG performs worst across all directions, indicating structural dissimilarity. In corn, HL and NC show strong within- and cross-region performance, while PG again fails to generalize (e.g., PG$\rightarrow$EU: 48.36). MMST-ViT exhibits degraded and highly variable performance, especially in cross-region settings (e.g., HL$\rightarrow$NGP: 25.01; PG$\rightarrow$HL: 64.08), suggesting poor transferability and possible overfitting. Table~\ref{tab:crop-model-performance} summarizes results across three OOD cases as defined in Table~\ref{tab:ood_scenarios}. GNN-RNN consistently achieves lower absolute RMSE across both corn and soybean predictions, even under OOD settings. This makes it a stronger candidate for deployment where minimizing prediction error is critical. However, the performance gap is more variable for GNN-RNN, particularly in harder OOD cases—indicating higher sensitivity to distribution shift. MMST-ViT, while slightly less accurate overall, exhibits more stable performance gaps across regions and crops.

\begin{table}[htbp]
  \caption{OOD vs. same-region RMSE (bu/acre) across crops, models, and scenarios. The scenario cases are detailed in Table~\ref{tab:ood_scenarios}.}
  \label{tab:crop-model-performance}
  \centering
  \scriptsize
  \begin{tabular}{lllccc}
    \toprule
    Crop & Model & Scenario & \makecell{RMSE(Diff region\\year-ahead)}& \makecell{RMSE\\(same-region\\year-ahead)} & \makecell{Performance \\Gap (\%)} \\
    \midrule
    \multirow{3}{*}{Soybean} & \multirow{3}{*}{MMST-ViT}
      & Case 1 & 9.04 & 8.93 & 1.23 \\
      & & Case 2 & 11.63 & 10.18 & 14.24 \\
      & & Case 3 & 12.19 & 11.33 & 7.59 \\
    \midrule
    \multirow{3}{*}{Corn} & \multirow{3}{*}{MMST-ViT}
      & Case 1 & 30.92 & 26.06 & 18.65 \\
      & & Case 2 & 34.93 & 33.37 & 4.67 \\
      & & Case 3 & 50.26 & 43.61 & 15.25 \\
    \midrule
    \multirow{3}{*}{Soybean} & \multirow{3}{*}{GNN-RNN}
      & Case 1 & 6.92 & 5.46 & 26.75 \\
      & & Case 2 & 9.75 & 6.15 & 58.53 \\
      & & Case 3 & 11.11 & 7.11 & 56.20 \\
    \midrule
    \multirow{3}{*}{Corn} & \multirow{3}{*}{GNN-RNN}
      & Case 1 & 27.62 & 26.69 & 3.48 \\
      & & Case 2 & 27.75 & 22.67 & 22.40 \\
      & & Case 3 & 32.07 & 21.65 & 48.13 \\
    \bottomrule
  \end{tabular}
\end{table}

\section{Discussion}

While CropNet provides the first large-scale multi-modal benchmark for U.S. county-level yield prediction, key limitations remain. Only 4 of Sentinel-2’s 12 spectral bands are used, excluding red-edge bands critical for early vegetation stress detection \citep{krisp2021rededge}. Imagery is Level-1C (uncorrected) \citep{topping2019atmospheric}, and spatial aggregation to $9\,\text{km} \times 9\,\text{km}$ grids erases field-level variability. Grid coverage per county varies greatly (5–130+), biasing learning toward large counties and degrading cross-region robustness. More uniform resolution sources like MODIS (1 km) could address some of these gaps. 

Across 1,200 counties, GNN-RNN showed better generalization than MMST-ViT, retaining positive correlation under USDA region shifts. MMST-ViT performed well in-domain but degraded sharply under OOD, revealing reliance on regional memorization. PG was consistently hardest to predict due to semi-arid climate, unmodeled irrigation, internal heterogeneity, sparse USDA labels, and missed stress signals due to omitted red-edge bands. This aligns with prior findings that temperature anomalies—not precipitation—drive global yield variation \citep{iizumi2020local}, highlighting the impact of lost local variability. The lack of comparisons to process-based baselines like DSSAT or APSIM \citep{lobell2015scalable} limits broader relevance. 

Finally, persistent underperformance in vulnerable, irrigated regions like PG raises equity concerns: if AI tools are more accurate in well-resourced rain-fed zones, they risk worsening existing agricultural disparities. Improving generalization through additional covariates, region-aware normalization, domain-adversarial methods, and hybrid physical–ML modeling is vital for both performance and fairness.

\section{Conclusion}

We present the first large-scale evaluation of deep learning models for crop yield prediction under realistic out-of-distribution (OOD) conditions. Our results show that GNN-RNN offers stronger cross-region generalization and is over 100× more compute resource efficient than MMST-ViT (Table ~\ref{tab:training_time_comparison}) making it more viable for sustainable deployment. MMST-ViT performs well in-domain but fails to generalize beyond the original four states, underscoring the importance of regionally diverse benchmarks. Both models struggle in structurally distinct zones like Prairie Gateway, where OOD performance gaps exceed 50\%. These findings reveal that spatial-temporal alignment—not just model complexity or data scale—is key to generalization. As climate change disrupts historical patterns, our work stresses the need for transparent OOD protocols to ensure robust and equitable agricultural forecasting.

\medskip

\bibliography{references}    

\end{document}